# Building the Language Resource for a Cebuano-Filipino Neural Machine Translation System

Kristine Mae M. Adlaon[1,2], Nelson Marcos[1]
[1]College of Computer Studies, De La Salle University, Taft Avenue, Manila, Philippines
[2]ITE Program, University of the Immaculate Conception, Father Selga St., Davao City, Philippines
[1]{kristine_adlaon},{nelson.marcos}@dlsu.edu.ph, [2]{kadlaon}@uic.edu.ph

## ABSTRACT

Parallel corpus is a critical resource in machine learning based translation. The task of collecting, extracting, and aligning texts in order to build an acceptable corpus for doing translation is very tedious most especially for low-resource languages. In this paper, we present the efforts made to build a parallel corpus for Cebuano and Filipino from two different domains: biblical texts and the web. For the biblical resource, subword unit translation for verbs and copy-able approach for nouns were applied to correct inconsistencies in translation. This correction mechanism was applied as a preprocessing technique. On the other hand, for Wikipedia being the main web resource, commonly occurring topic segments were extracted from both the source and the target languages. These observed topic segments are unique in 4 different categories. The identification of these topic segments may be used for automatic extraction of sentences. A Recurrent Neural Network was used to implement the translation using OpenNMT sequence modeling tool in TensorFlow. The two different corpora were then evaluated by using them as two separate inputs in the neural network. Results have shown a difference in BLEU score in both corpora.

## CCS Concept:
Computing Methodologies → Machine Translation

## 1. INTRODUCTION

Machine translation (MT) in recent years had gained so much interest to many researchers because of its fascinating use in different tasks. Both statistical and neural machine translation approaches are now paving its way towards improving MT in different languages [3],[4],[8]. The Philippines, being a linguistically diverse country could greatly benefit from having an effective MT system as this could help improve in the promotion of educational and cultural development agenda of the country. There have been several efforts already in machine translation for Philippine languages such as that of [5], [7], [9] but no known work had focused on purely Philippine native language alone using the state-of-the-art approaches for MT.

Knowing that both statistical and neural machine translation approaches are highly reliant on the availability of large amounts of data and are known to poorly perform on low-resource settings, this has been then the main motivation of this research work. In this paper, an approach in building the language resource for Cebuano and Filipino (formerly Tagalog) is presented. These two languages are the top most spoken languages in the Philippines (PH). The source of data for this study is parts of the Bible taken from the work of [1], and Philippine geographical information from Wikipedia. We also present in this paper the intensive labor of examining and correcting the source data to come up with a more reliable parallel corpus. We then compare the BLEU scores of the machine translation of the two sources to evaluate whether there is an effect in the quality of the translation given different domain of source using a unidirectional implementation of a recurrent neural network.

## 2. RELATED WORK

To date, efforts in building the language resource for the Philippines are very minimal much more in the curation of parallel corpora for machine translation. The only Cebuano to Filipino (Tagalog) MT effort in the Philippines is that of Fat (2007) who worked on a bilingual machine translation system designed for Tagalog and Cebuano. It exploits structural similarities of the Philippine languages Tagalog and Cebuano, and handles the free word order languages. It translates at the syntactic level only and uses Tagalog to Cebuano dictionary as the dataset. It does not employ morphological analysis in the system. A more recent work used a statistical machine translation (SMT) approach for a bidirectional Filipino to English MT named as the ASEANMT-Phil. The system has experimented on different settings producing the BLEU score of 32.71 for Filipino to English and 31.15 for English to Filipino [9].

The approach that was used in this study for inconsistency correction is greatly inspired by the work of [10]. Their work introduced a simpler and more effective approach, making the NMT model capable of open-vocabulary translation by encoding rare and unknown words as sequences of subword units. Several attempts for building a parallel corpus from the Web have already been conducted. The work of Ramesh and Sankaranarayanan [11] have used an end-to-end Siamese bidirectional recurrent neural network to generate parallel sentences from comparable multilingual articles in Wikipedia. Their study has shown that using the harvested dataset improved BLEU scores on both NMT and phrase-based SMT systems for the low-resource language pairs: English– Hindi and English–Tamil.

## 3. APPROACH

In this section, we describe the entire pipeline that this research went through as depicted in Figure 1.

### 3.1 Extraction of the Source Data

The Philippines has very few linguistic resources most especially for parallel corpora. In the past, efforts were made to build the Philippine language resource but were much more focused on Filipino only or English – Filipino for bilingual corpora. To the best of our knowledge at present, no known work for building the parallel corpus for Cebuano-Filipino is currently being done in the Philippines. In this subsection we describe the nature of the source dataset.

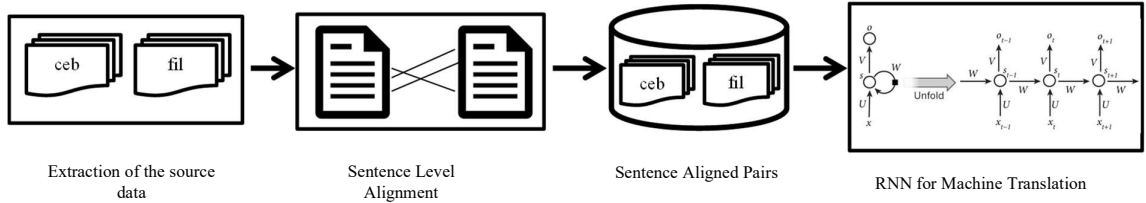

Figure 1: Architecture for the building of the parallel corpus showing four different steps: (a) Extraction of the source data, (b) Sentence Level Alignment, (c) Sentence Aligned Pairs, (d) unidirectional RNN for Machine Translation

Table 1: General information of the dataset

| Language | Cebuano | Filipino |
|---|---|---|
| Family | Austronesian | Austronesian |
| Genus | Malayo-Polynesian | Malayo-Polynesian |
| Subgenus | Philippines | Philippines |
| No. of Speakers | 15,800,000 | 23,900,000 |
| Parts (New or Old Testament) | Complete | Complete |

**Bible:** The dataset was downloaded from the web Bible[1] that contains a monolingual corpus of different languages created from translations of the Bible which is publicly available. Although it is stated in the source that the corpus's Book, Chapter, and Verse indices are already aligned at a sentence level, inconsistencies in the sentence level translation were still found. Thus, a manual correction (discuss in the next section) was done to address this. The corpus is in XML format. Table 1 shows the general information of the dataset.

Data cleaning and transformation was performed in the dataset. This stage is important in order to transform the data into a format that can be used for processing. We converted the dataset from XML (.xml) file to a text (.txt) file then removed all xml tags and non-printable ASCII characters. We also removed all repetitive sentences that were found mostly in the Cebuano dataset. For the purpose of this research, to check as to what will be the performance of this dataset to translation we only included the Genesis book. After performing the cleaning and transformation of the dataset, a total of 6,510 sentence pairs were generated. Table 2 shows example sentence pairs where Cebuano is the source sentence and Filipino is the target sentence.

**Wikipedia:** Wikipedia is an online collaborative encyclopedia available in a wide variety of languages. Wikipedia has aligned article pairs that may range from being almost completely parallel to containing almost no parallel sentences. To extract articles from Wikipedia we used a python library Beautiful Soup[2]. The URL to be used for pulling data is unique per language following the format shown in Figure 2.

Table 2: Cebuano - Filipino Sentence Pairs from the Bible

| Cebuano | Tagalog |
|---|---|
| Ug mitubag si jose kang faraon : ang damgo ni faraon usa lamang : ang dios nagpahayag kang faraon sa haduol na niyang pagabuhaton . | At sinabi ni jose kay faraon , ang panaginip ni faraon ay iisa; ang gagawin ng dios ay ipinahayag kay faraon : |
| Ang pito ka mga vaca nga maanindot mao ang pito ka tuig ; ug ang mga uhay nga maanindot mao ang pito ka tuig : ang damgo usa lamang . | Ang pitong bakang mabubuti ay pitong taon ; at ang pitong uhay na mabubuti ay pitong taon ; ang panaginip ay iisa . |

Language code *ceb* is used for Cebuano while *tl* is used for Filipino. They both share the same topic as indicated in the *wiki_article*.

```
wiki_article = "Metro_Manila"
lang_code = ceb or tl

url = 'https://[lang_code]
.wikipedia.org/wiki/'+wiki_article
```

Figure 2: URL format used in scraping

We manually identified categories to be extracted from Wikipedia. These categories were chosen since we have seen that articles in these categories have available data for both languages. The categories are PH regions, PH provinces, PH cities, and PH tourism attractions with 17, 82, 114, and 22 counts respectively. Regions, provinces, and cities are also called administrative divisions of the Philippines, often lumped together as Local Government Unit (LGU). Shown in Table 3 are sample snippets of the extracted data.

To match with the number of sentence pairs extracted from the Bible, we have also extracted a total of 6,510 sentence pairs for Wiki articles. We were able to obtain 1537, 1929, 2044, and 1000 sentences for regions, provinces, cities, and tourism attractions respectively.

---

[1] http://christos-c.com/bible/
[2] https://www.crummy.com/software/BeautifulSoup/bs4/doc/

## 3.2 Sentence Level Alignment

In this section, we discuss the strategies that we applied for both the Bible and Wikipedia texts to come up with a sentence aligned parallel corpus. Most of these strategies were applied to address inconsistences seen in the extracted data.

Table 3: Snippet of the texts extracted from Wikipedia.

| Category | A snippet of the extracted article |
|---|---|
| Regions | Ang Caraga maoy usa sa mga rehiyon sa Pilipinas nga makita sa pulo sa Mindanao. |
| | Ang Caraga ay isang rehiyon ng Pilipinas na matatagpuan sa hilagang silangang bahagi ng pulo ng Mindanao. |
| Provinces | Ang Abra usa ka lalawigan sa Pilipinas sa Administratibong Rehiyon sa Cordillera sa Luzon. |
| | Ang Abra (Ilokano:Probinsiya ti Abra) ay isang lalawigan ng Pilipinas na matagpuan sa Cordillera Administrative Region sa Luzon. |
| Tourism Attractions | Ang Malolos primera klaseng dakbayan nahimutang sa probensya sa Bulacan, Pilipinas. |
| | Ang Lungsod ng Malolos ay isang unang uring lungsod sa Pilipinas sa lalawigan ng Bulacan. |

**Bible:** In this source, we have observed that inconsistences in translation were mostly prevalent on nouns and verbs. We have seen inconsistences in the translation of names of people in the Bible from source to target which we think are very critical to address since the Bible has a lot of important biblical names. Shown in Table 4 are samples that were seen as inconsistent translations for most mentioned names in the source:

Table 4: Sample names extracted from the dataset with high frequency

| Names | Possible Translations | Replaced with | # of Instances |
|---|---|---|---|
| jehova | jehova, panginoon, dios none | jehova | 1580 |
| dios | jehova, panginoon, dios none | dios | 622 |
| israel | katilingban, israelita, israelihanon, none | israel | 665 |
| moises | moises, siya, kaniya, none | moises | 651 |
| aaron | aarong | aaron | 292 |

It can be observed in Table 4, that both the words *jehova* and *dios* can be used to translate one from the other. There are instances in the dataset that the word *jehova* from the source is referred to as *dios* or *panginoon*. The *none* tag in the possible translation column is a case where in there is no equivalent translation of the word in the target language given a sentence pair. It can be noticed as well that most of the possible translations for the name *moises* are pronouns such as the words *siya, kaniya*.

Word translation was performed for most frequently occurring names that were inconsistent translated. We applied the copy-able approach for word translation in names. Example, for the word *dios* it was found out that the most frequent translation of the word *dios* is also *dios*, therefore any instance that the word *dios* is found in the source language and is translated as *panginoon* or *jehova*; it will be replaced with the word *dios*. If none, the word *dios* will be inserted in the target language.

Table 5: Sample list of verbs and its possible translation

| Verbs | Possible Translations | # of Instances |
|---|---|---|
| ngadto | pumaroon, bumaba, pasubasob, none, magsisiyaon, nagsibaba, napasa, mula, paroon, yumaon, papasasa | 541 |
| sumala | ayon, gaya ng, kung paanong | 273 |
| uban | kasalamuha, none | 332 |

Aside from names, most frequently occurring verbs found in the dataset were examined. Table 5 shows 5 of the most frequently occurring verbs. Subword unit translation [3]. was performed in these words. Example, the word *ngadto* has 11 possible translations. After inspection it was found out that the most frequently used translation of the word *ngadto* is *paroon*. After knowing that the translation is pa*roon*, all occurrences of the word *ngadto* in the source language was replaced with the word *paroon* in the target language.

**Wikipedia:** In this resource, we observed that majority of the articles has more number of sentences available in Filipino than in Cebuano. We manually examined each of the text files and exhaustively look up for sentence pairs that are translations of each other. We then observed that some of the sentence pairs for each category exhibits common features.

For regions, the word sequence *maoy rehiyon sa Pilipinas* in Cebuano, *ay isang rehiyon sa Pilipinas* in Filipino, *is a region in the Philippines* in English appeared 205 times or 13% of the sentence pairs under this category contains this word sequence. Examples of these sentences are:

1. Ang Lupot sa Cagayan maoy rehiyon sa Pilipinas.
2. Ang Administratibong Rehiyon sa Cordillera maoy rehiyon sa Pilipinas.
3. Ang Calabarzon maoy rehiyon sa Pilipinas nga gilangkuban sa mga musunod nga lalawigan: Cavite, Laguna, Batangas, Rizal at Quezon.

Other word sequence commonly found under this category is *gilangkuban kini sa [count] ka lalawigan* followed by a colon (:)

For cities, same as in provinces the word feature *ceb::nabahin or fil::nahahati* frequently occurs in the extracted texts. While provinces are composed of cities, cities on the other hand are composed of *barangays*.

## 4. EXPERIMENT RESULTS

In this section, results of training Cebuano to Filipino sentence pairs in a unidirectional Recurrent Neural Network (RNN) with and without corrections to the corpus are shown. Three experiments were performed: (1) the RNN model was fed with sentence pairs without subword translation, (2) the RNN model was fed with sentence pairs with subword translation, and (3) the RNN model was trained using the Wikipedia sentence pairs.

Table 6: Comparison of Translated Sentences with and without subword translation

| SP | Reference | Result without Subword Translation | BLEU Score | Result with Subword Translation | BLEU Score |
|---|---|---|---|---|---|
| 1 | at sinabi ng dios , *sibulan* ang lupang *damo* , *pananim* na *nagkakabinhi* , at *punong* kahoy na *namumunga ayon* sa *kaniyang pagkakahoy* , na *taglay* ang *kaniyang* binhi sa *ibabaw* ng lupa , at *nagkagayon* . | at sinabi ng dios , *purihin* ang l upa , at *iniibig* ang lupa , na *tin implahan* ng *kapahingahan* , at *tatalab* sa *kaniya* ang *mga* kahoy na *iyon* , na *parang bunga* ng *kanilang* binhi : at *magiging banal* sa *kaniya* . | 16.52 | at sinabi ng dios , *mapayapa nawa* ang *lupain* , at *iginuhit* sa *iyo* , ng *kasagutan* at ng *laryo* , na *mga* kahoy na *itinutubo* ng kahoy sa lupain ; at *siya'y naging parang pakinabang* sa ibabaw ng lupa . | 18.97 |
| 2 | at ang *lupa* ay *sinibulan* ng *damo* , *pananim* na *nagkakabinhi* , ayon sa *kaniyang pagkapananim* , at ng *punong*kahoy na *namumunga* , na *taglay* ang *kaniyang* binhi , ayon sa *kaniyang pagkakahoy* , at nakita ng dios na mabuti | at ang *lupain* ay *mapasuko* , at *kumapal* na *parang* binhi ng *pakikiapid* , na *gaya* ng *kanilang ama* : at ang *mga* kahoy na *cedro* , ay ayon sa *kanilang kapanganakan* , at nak ita ng dios na mabuti . | 24.08 | at ang *lupain* ay *mapasuko* at *nilakad* , , na *mainam* na *harina* , at ang binhi ng *pakikiapid* , at ang *mga* kahoy na *itinutubo* ng kahoy na *cedro* , ay *magtataglay* ng *kanilang ulo* at ng dios na mabuti . | 15.05 |
| 3 | at *nagkahapon* at *nagkaumaga* ang *ikatlong* araw . | at *may malakas na hiyawan sa* araw *na yaon* . | 5.52 | at nagkahapon at nagkaumaga ang ikatlong araw . | 100.00 |

### 4.1 Without Subword Translation (Bible)

After splitting the corpus, training was performed. Logarithmic loss values generated during the training to measure the performance of the model on how far the source values are from the target indicated values were extracted. The ideal loss value is 0. Training was stop at step 25000 (loss = 1.39) when loss was consistently below 2 starting step 21000 (loss = 2). After the training, translation was evaluated using the test cases via the BLEU metric. Result of the evaluation is a BLEU score indicating that the higher the BLEU score the better the translation.

The model achieved a BLEU score of **20.01** for Cebuano to Filipino translation without the word and subword translation. This score indicates a somewhat understandable translation. Table 4 shows a sample translated sentences taken from the test set. The words that are strongly emphasized and italicized are the words not found in the translation.

Translations seem to have a much better BLEU score result on longer sentences compared to short sentences. This is because unigram correct translation is much higher on longer sentences. Although this is not true in all sentence pairs seen on the test set. An example would be the translation for this source sentence:

  *ug ang babaye mitubag sa bitin : makakaon kami sa bunga sa mga kahoy sa tanaman :*

The correct translation should be:

  *at sinabi ng babae sa ahas , sa bunga ng mga punong kahoy sa halamanan ay makakakain kami*

The machine translated it as:

  *at sinabi ng babae sa ahas , kami ay tumakas sa bunga ng mga punong kahoy ,*

This sentence pair got a high BLEU score of 58.07. In the translation table result shown in Table 6, it can be observed that in sentence pair 2 there is an insertion of the word *cedro* (a noun based on context) which is never seen in the source sentence. When the word *cedro* was searched in the whole training set, we found that the word *cedro* is a name of a tree which was always mentioned in the training set as *kahoy na cedro*. Looking at the sentence pair 2 in the test set, where punong kahoy was mentioned, the machine translated it as *kahoy na cedro*.

### 4.2 With Subword Unit Translation (Bible)

After performing subword translation as discussed in section 3.2, logarithmic training loss values generated during the training to measure the performance of the model on how far the source values are from the target indicated values were extracted. The ideal loss value is 0. Training was stop at step 9000 (loss = 1.22) when loss was consistently below 2.

The training, translation was evaluated using the test cases via the BLEU metric. The model achieved a slightly higher BLEU score of **22.87** for Cebuano to Tagalog translation with subword translation. Table 6 shows that the impact of the correction (Subword translation) is greatly seen on sentence pair number 3. A perfect translation of the reference sentence was outputted. On the other hand it did not performed well on sentence pair number 2. The BLEU score got even lower.

### 4.3 Wikipedia

After splitting the corpus, training was performed. Just like the network parameter setup used in the Bible, we also used the same for Wikipedia. Logarithmic loss values generated during the training to measure the performance of the model on how far the source values are from the target indicated values were extracted. The ideal loss value is 0.

Training was stop when loss was consistently below 2. After the training, translation was evaluated using the test cases via the BLEU metric. Result of the evaluation is a BLEU score indicating that the higher the BLEU score the better the translation.

The model achieved a BLEU score of **27.36** for Cebuano to Filipino translation. An increase of **4.46** was seen, slightly higher compared to that of the Bible dataset with subword unit translation. We observed that the model performed best in translating the first n-gram feature of the sentence such as the following sentences:

Table 7: Observed correct translation of the first word of the sentence

| Reference | Translation |
|---|---|
| **Ang simbahan ay** mayaman **na** pinagkalooban, **na** may masarap **na** retablo, pulpito, lectern at koro-kuwadra . | **Ang simbahan ay** napinsala **na** isang mapanganib **na** katibayan kung nakukuha ng mga pwersa ng kaaway , . |
| **Ang mga espesyal na** interes **ay** ang serye **ng** crypto-collateral chapel **na** nagsasapaw **sa magkabilang panig ng** nave . | **Ang espesyal na** konstruksyon **ay** ginawa **ng mga** durog **na** bato **na** nakatuon **sa magkabila ng panig ng** pasukan . |

The model performed poorly on translation numbers such as dates, counts, and etc.

## 5. Conclusion

In this paper, a never before applied approach for language translation was applied for a Cebuano to Tagalog translator. A Recurrent Neural Network was used to implement the translator using OpenNMT framework. The performance of the translation was evaluated using the BLEU Score metric. Two different types of data were used in the experiments: religious (Bible) and data from Web (Wikipedia). To address inconsistencies of translation in the source dataset, a subword unit translation was performed. This correction results to an increased BLEU score of 22.87 from 20.01.

Although, there was a slight increase in the translation performce, this still indicates a somehow understandable translation. For the Wikipedia dataset, the model performed better than that of the Bible with a slightly higher score of 4.46. We observed that the model performed best in translating the first n-gram feature of the sentence and performed poorly on numeric features.

For future work, it is recommended to increase the number of training pairs to improve the translation performance. We also recommmend to leverage on the use of the identified topic segments or word sequences in Wikipedia to automatically extract parallel sentences.

## 6. Acknowledgments

We would like to thank De La Salle University and University of the Immaculate Conception for the financial support. Also, the Philippine Commission on Higher Education for the scholarship granted to one of the researchers.